\documentclass[dvipsnames,format=sigconf,anonymous=false,review=false]{acmart}

\AtBeginDocument{%
  }

\copyrightyear{2026}
\acmYear{2026}
\setcopyright{cc}
\setcctype{by}
\acmConference[GECCO '26]{Genetic and Evolutionary Computation Conference}{July 13--17, 2026}{San Jose, Costa Rica}
\acmBooktitle{Genetic and Evolutionary Computation Conference (GECCO '26), July 13--17, 2026, San Jose, Costa Rica}
\acmDOI{10.1145/3795095.3805085}
\acmISBN{979-8-4007-2487-9/2026/07}

\usepackage[linesnumbered,ruled,vlined]{algorithm2e}
\usepackage{soul}

\begin{document}
\sloppy


\title{BACE: LLM-based Code Generation through Bayesian Anchored Co-Evolution of Code and Test Populations}

\author{Kaushitha Silva}
\affiliation{%
  \institution{WSO2}
  \city{Santa Clara}
  \state{CA}
  \country{USA}
}
\email{kaushitha@wso2.com}
\orcid{0009-0005-0098-7815}

\author{Srinath Perera}
\affiliation{%
  \institution{WSO2}
  \city{Santa Clara}
  \state{CA}
  \country{USA}}
\email{srinath@wso2.com}
\orcid{0000-0002-4457-903X}


\begin{abstract}
Large Language Models (LLMs) have demonstrated impressive capabilities in code generation. While an interactive feedback loop can improve performance, writing effective tests is a non-trivial task. Early multi-agent frameworks, such as AgentCoder, automated this process but relied on generated tests as absolute ground truth. This approach is fragile: incorrect code frequently passes faulty or trivial tests, while valid solutions are often degraded to satisfy incorrect assertions. Addressing this limitation, newer methods have largely abandoned test generation in favor of planning and reasoning based on examples. We argue, however, that generated tests remain a valuable signal if we model them as noisy sensors guided by Bayesian updates. To this end, we introduce \textbf{BACE (Bayesian Anchored Co-Evolution)}, a framework that reformulates synthesis as a Bayesian co-evolutionary process where code and test populations are evolved, guided by belief distributions that are reciprocally updated based on noisy interaction evidence. By anchoring this search on minimal public examples, BACE prevents the co-evolutionary drift typical of self-validating loops. Extensive evaluations on LiveCodeBench v6 (post-March 2025) reveal that BACE achieves state-of-the-art performance across both proprietary models and open-weight models at 7B and 120B scales.
\end{abstract}




\maketitle

\section{Introduction}
\label{sec:intro}

The automated synthesis of verified software from natural language is the central pursuit of modern software engineering. While Large Language Models (LLMs) have achieved unprecedented milestones in code generation, a critical bottleneck remains: LLMs frequently synthesize solutions containing subtle logical errors that bypass traditional open-loop generation~\cite{wang_towards_2025}. To resolve this, closed-loop developer--tester paradigms (e.g., AgentCoder~\cite{huang_agentcoder_2024}) have used iteratively repairing code based on test execution feedback, achieving substantial performance gains over single-shot prompting.

However, the efficacy of this paradigm is fundamentally limited by the reliability of the test generation agent. As acknowledged by the authors \cite{huang_agentcoder_2024}, incorrect test cases produce problematic feedback, misleading the programmer agent. The reliance on generated tests creates a fragile loop where "false positives" occur because incorrect code satisfies hallucinated or trivial tests, and valid solutions are frequently degraded by "false negatives" resulting from faulty assertions. As we will see in the experiment section, AgentCoder underperformed even the Direct prompting baseline with GPT-OSS-120b.  

\begin{figure}[t]
    \centering
    \includegraphics[width=\columnwidth]{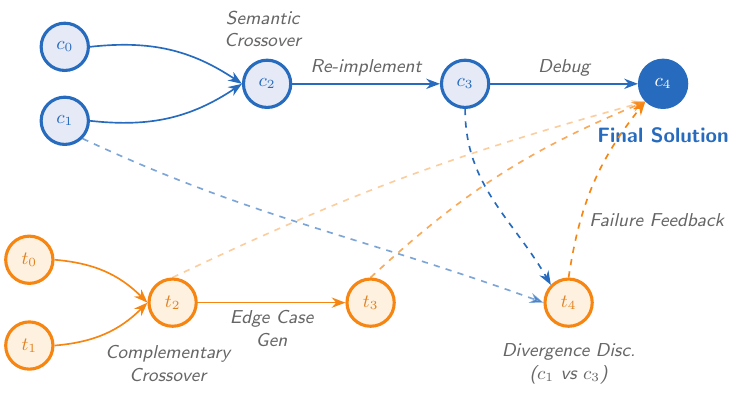}
    \Description{A flowchart diagram illustrating the co-evolution of code candidates (blue nodes, top) and test cases (orange nodes, bottom). 
    In the code evolution track, candidates $c_0$ and $c_1$ combine via "Semantic Crossover" to form $c_2$. $c_2$ evolves via "Re-implement" into $c_3$. $c_3$ evolves via "Debug" into the dark blue node $c_4$, labeled "Final Solution."
    In the test evolution track, tests $t_0$ and $t_1$ combine via "Complementary Crossover" to form $t_2$. $t_2$ evolves via "Edge Case Gen" into $t_3$.
    Interactions between tracks are shown with dashed arrows: Code nodes $c_1$ and $c_3$ point to a test node $t_4$, labeled "Divergence Disc. ($c_1$ vs $c_3$)." Finally, feedback arrows labeled "Failure Feedback" point from test nodes $t_2$ and $t_4$ to the final debugging step that produces $c_4$.}
    \caption{\textbf{Ancestral Lineage of a Solution.} An illustrative visualization of the BACE co-evolutionary process, tracing the genesis of a candidate solution ($c_4$)}
    \label{fig:evolutionary_lineage}

\end{figure}

Due to this unreliability, the most recent state-of-the-art multi-agent frameworks, such as MapCoder~\cite{islam_mapcoder_2024} and CodeSIM~\cite{islam_codesim_2025} have abandoned test generation. These systems instead focus on pure reasoning (planning, simulating..etc), effectively conceding that generated tests are too unreliable to serve as good signal to improve code generation. We contend that generated tests remain valuable if we develop methods to leverage them despite their lack of ground truth. The central research question effectively becomes: \textbf{How can a synthesis system converge on a correct solution when the instrument of measurement---the test suite---is itself an unreliable source of truth?}

To address this, we introduce \textbf{BACE (Bayesian Anchored Co-Evolution)}. Our approach begins by maintaining a population of code and tests rather than single instances. This population-based structure partially mitigates the risk of bad feedback: even if a valid solution is degraded by a faulty test in one instance, the broader population may ensure that other valid genetic lineages survive, preventing the immediate loss of correct logic. 

However, populations alone are insufficient. If fitness is calculated under the assumption that generated tests are ground truth, the population will simply converge on the errors, leading to co-evolutionary drift. To solve this, BACE abandons the deterministic view of testing. We model fitness as a belief in an individual's correctness based on a Bayesian formulation. We treat execution results as noisy signals rather than binary gates, utilizing these observations to reciprocally update the belief distributions of both the code and test populations.
Finally, to prevent the entire system from drifting into self-validating loops, we \textbf{``Anchor''} the code solution on the minimal set of 1--3 input/output samples provided in the problem specification. Figure~\ref{fig:evolutionary_lineage} illustrates how candidate solutions are formed through various evolutionary operations. 

Averaged over three runs on LiveCodeBench v6 (post-March 2025)~\cite{jain_livecodebench_2024}, BACE outperforms leading multi-agent frameworks, surpassing CodeSIM by absolute margins of \textbf{2.5\%} with GPT-5-Mini, \textbf{5.4\%} with Qwen2.5-Coder-7b, and \textbf{5.0\%} with GPT-OSS-120b.

The major contributions of this paper are as follows:
\begin{itemize}
    \item \textbf{Bayesian Co-Evolutionary Framework}: We reformulate code synthesis as a co-evolutionary process where code and test populations reciprocally evolve based on belief distributions updated via noisy interaction evidence.
    \item \textbf{Belief Anchoring Mechanism}: We introduce an ``Anchoring'' mechanism that conditions belief updates on minimal public examples, mitigating co-evolutionary drift.
    \item \textbf{Behavioral Diversity Retention}: We enforce population diversity through two novel strategies: a behavioral-based elitism policy that preserves individuals with unique behavioral vector, and the strategic use of differential testing to discover diversity.
    \item \textbf{State-of-the-Art Performance}: We demonstrate that BACE achieves superior performance on LiveCodeBench v6 (post-March 2025) compared to leading multi-agent frameworks in both state-of-the-art proprietary models and open-weight models.
\end{itemize}

\section{Related Work}
\subsection{Pre-LLM: Program Synthesis}
Program synthesis is the automatic construction of an executable program that satisfies a given high-level specification~\cite{manna_toward_1971}. One prominent paradigm, Search-Based Software Engineering (SBSE), reformulates synthesis as a search-based optimization problem. A foundational technique in this domain is Genetic Programming (GP), popularized by Koza~\cite{koza_genetic_1994}, which adapts genetic algorithms to evolve executable structures. To expand representational flexibility beyond standard GP, Grammatical Evolution (GE)~\cite{goos_grammatical_1998, mcdermott_grammar_2017} maps linear genotypes to source code via context-free grammars, while approaches like PushGP~\cite{spector_autoconstructive_2001} utilize stack-based execution architectures to naturally handle multiple data types and complex control flows. 


Standard GP is prone to premature convergence. Hillis~\cite{hillis_co-evolving_1990} addressed this via `host-parasite' coevolution, demonstrating that evolving adversarial test cases alongside solutions can force the search out of local optima. Arcuri and Yao~\cite{arcuri_coevolving_2007} applied this to program synthesis, co-evolving programs against unit tests. Unlike standard GP's static test suites, their method evolves adversarial tests to actively expose faults, creating a competitive arms race that enforces strict adherence to the specification. However, these co-evolutionary methods typically rely on formal specifications to validate correctness. 

\subsection{Large Language Models for Code Generation}

The emergence of Large Language Models (LLMs) has significantly advanced NL-based program synthesis. Hernandez et al.~\cite{hernandez_gp_2025} shows there are ``no clear winners'' among GP and LLMs when evaluated against PSB2~\cite{helmuth_psb2_2021}, a benchmark suite often utilized for GP evaluations. Conversely, on LLM-centric benchmarks such as HumanEval, Custode et al.~\cite{custode_comparing_2024} found that LLMs significantly outperform GE. While contemporary models now achieve near-perfect scores on HumanEval~\cite{chen_evaluating_2021}, these metrics are increasingly suspect due to data contamination. To mitigate this, Jain et al.~\cite{jain_livecodebench_2024} introduced Live-CodeBench to evaluate models on uncontaminated, post-training problems, revealing substantial performance drops. These findings, consistent with observations by Wang et al.~\cite{wang_towards_2025}, indicate that while LLMs demonstrate robust coding capabilities, zero-shot inference alone remains insufficient to guarantee correctness on novel tasks.

\subsection{Enhancing LLM-Driven Code Generation}
To enhance zero-shot capabilities, research has focused on optimizing inference strategies. Initial ``open-loop'' methods, such as Chain-of-Thought (CoT)~\cite{wei_chain--thought_2023}, Structured SCoT (SCoT)~\cite{li_structured_2023}, and Self-Planning \cite{jiang_self-planning_2024}, encourage models to decompose specifications into intermediate reasoning steps. While these strategies were effective, subsequent research established that integrating actual execution feedback improves performance compared to purely open-loop reasoning. Consequently, ``closed-loop'' approaches like Reflexion~\cite{shinn_reflexion_2023}, CodeCoT~\cite{huang_codecot_2024}, INTERVENOR~\cite{wang_intervenor_2024}, and Self-Debugging~\cite{chen_teaching_2023} incorporated iterative feedback from execution traces, either by generating their own tests or by using publicly available tests. These were then scaled up to multi-agent architectures that simulate software development workflows. For instance, AgentCoder~\cite{huang_agentcoder_2024} coordinates programmer, test designer, and executor agents to iteratively refine code through generated unit tests. Similarly, DebateCoder~\cite{chen_debatecoder_2025} and CodeCoR~\cite{pan_codecor_2025} also used automated test cases.

A critical bottleneck in these feedback loops is the reliability of the validation mechanism itself. As noted by Huang et al~\cite{huang_agentcoder_2024}, incorrect oracles provide misleading signals that degrade performance. To mitigate this, frameworks like MapCoder~\cite{islam_mapcoder_2024} and CodeSIM~\cite{islam_codesim_2025} employ multi-agent collaboration strategies that bypass test generation entirely, focusing instead on reasoning based on input/output examples. Alternatively, CodeT~\cite{chen_codet_2022} leverages dual execution agreement by generating both solution candidates and test cases. It identifies consensus sets where clusters of solutions pass the same subset of tests, selecting the final solution from the set maximizing the combined size of code and test populations. This, however, fails if the wrong solution clusters are bigger.

The work most closely related to ours is CoCoEvo~\cite{li_cocoevo_2025}, which applies a co-evolutionary approach to code and test generation. CoCoEvo defines code fitness using a CodeT-style dual agreement confidence score, while test fitness is determined via a Pareto frontier of confidence and discrimination scores. Due to the lack of reported results on standard benchmarks and the unavailability of the source code, we are unable to benchmark CoCoEvo against the state of the art or include it as a baseline. 

Our approach distinguishes itself from these predecessors through three key mechanisms. First, unlike the stochastic discovery in Arcuri and Yao~\cite{arcuri_coevolving_2007}, BACE leverages LLMs to initialize plausible candidates, focusing evolution on refinement rather than discovery from scratch. Second, in contrast to AgentCoder~\cite{huang_agentcoder_2024}, which treats generated tests as absolute ground truth, BACE mitigates the risk of degrading valid solutions by maintaining a population and modeling generated tests as noisy sensors within a Bayesian framework and diversity retention via pass-fail behavioral vectors. Finally, while MapCoder~\cite{islam_mapcoder_2024} and CodeSIM~\cite{islam_codesim_2025} abandon test generation to avoid the inherent unreliability of synthesized oracles, we demonstrate that anchoring the belief effectively grounds the search, allowing the system to safely exploit the high-value signal from generated tests.
\section{Probabilistic Modeling of Code-Test Interactions}

To address the limitations of deterministic oracles while incorporating test feedback, we reformulate the co-evolutionary process within a Bayesian framework as follows. We build on the Bayesian evolutionary computation framework introduced by Zhang et al.~\cite{zhang_bayesian_1999} while extending it to accommodate \textit{co-evolution}.

\subsection{Latent Variables and Belief Notation}
We define a belief distribution representing our confidence in the correctness of code and tests relative to a problem specification $\mathcal{S}$. Specifically, let $\mathcal{C} = \{c_1, \dots, c_N\}$ represent a population of candidate solutions and $\mathcal{T} = \{t_1, \dots, t_M\}$ represent a population of tests. We associate each individual with a binary latent variable representing its ground-truth correctness:
\begin{itemize}
    \item $X_i \in \{0, 1\}$ denotes the correctness of code candidate $c_i$, where $1$ is correct.
    \item $Y_j \in \{0, 1\}$ denotes the validity of test case $t_j$, where $1$ is a valid test of $\mathcal{S}$.
\end{itemize}

We define the \textit{belief} in an individual as its posterior probability of correctness. For brevity, we denote:
\begin{equation}
    b(c_i) = P(X_i = 1) \quad \text{and} \quad b(t_j) = P(Y_j = 1)
\end{equation}

The interaction between $c_i$ and $t_j$ produces an outcome $d_{ij} \in \{0, 1\}$, where $1$ indicates a ``Pass'' and $0$ indicates a ``Fail''. These outcomes are aggregated into an observation matrix $\mathbf{D} \in \{0, 1\}^{N \times M}$, which serves as the primary data source for belief propagation.

\subsection{The Noisy Sensor Model}
\label{subsec:noisy_sensor}
When evaluating generated code against generated tests, we cannot interpret execution results as ground truth. In traditional testing, a passing result typically implies the code is correct. However, in our setting, a passing execution ($d_{ij}=1$) is an ambiguous signal: it does not necessarily imply that the code $c_i$ is correct, nor that the test $t_j$ is valid. A "Pass" can occur even if the code is buggy (e.g., if the test is trivial) or if the test is wrong (e.g., matching a wrong output).

To handle this uncertainty, BACE treats the execution outcome not as a definitive verdict, but as a ``noisy sensor'' observation conditioned on the binary latent variables. While we assume that correct code passing a valid test is deterministic ($P(d_{ij} = 1 \mid X_i=1, Y_j=1) = 1$), a "Pass" can also be generated in three degenerate scenarios. We quantify the likelihood of these misleading signals using three noise hyperparameters:\begin{enumerate}\item \textbf{False Pass} ($\alpha$): The probability that valid code passes a broken test ($P(d_{ij} = 1 \mid X_i=1, Y_j=0)$).\item \textbf{Accidental Pass} ($\beta$): The probability that incorrect code passes a valid test ($P(d_{ij} = 1 \mid X_i=0, Y_j=1)$).\item \textbf{Coincidental Pass} ($\gamma$): The probability that incorrect code passes a broken test ($P(d_{ij} = 1 \mid X_i=0, Y_j=0)$).\end{enumerate}

\subsection{Belief Updates}
Based on test execution results, we update the belief of an individual using the Bayesian update rule. The updates are performed in log-odds space ($\operatorname{logit}(p) = \ln \frac{p}{1-p}$) to ensure numerical stability and additive evidence aggregation. To prevent premature convergence in the early generations, especially when using large populations, we apply a learning rate $\eta \in (0, 1]$. For a code candidate $c_i$, the updated belief is defined as:
\begin{equation}
    \operatorname{logit}(b(c_i)) = \operatorname{logit}(b_{\text{prior}}(c_i)) + \eta \sum_{j} \Delta_{ij}^{\text{code}}
\end{equation}
where $\Delta_{ij}^{\text{code}}$ represents the Weight of Evidence (WoE) provided by test $t_j$. A symmetric update is applied to test beliefs $b(t_j)$ by aggregating evidence $\Delta_{ji}^{\text{test}}$ from the code population. 

We avoid double-counting evidence by excising observations $d_{ij}$ that have been previously accounted for in prior generations. (Refer the appendix for the more detailed derivation of the belief update equations)

\subsubsection{Code Belief Update}

For a code candidate $c_i$, the evidence $\Delta_{ij}^{\text{code}}$ is conditioned on the current belief of the test $b(t_j)$:
\begin{equation}
    \label{eq:code_woe}
    \Delta_{ij}^{\text{code}} = 
    \begin{cases} 
    \ln \left( \frac{b(t_j) + \alpha (1 - b(t_j))}{\beta b(t_j) + \gamma (1 - b(t_j))} \right) & \text{if } d_{ij}=1 \\
    \ln \left( \frac{(1-\alpha)(1 - b(t_j))}{(1-\beta)b(t_j) + (1-\gamma)(1 - b(t_j))} \right) & \text{if } d_{ij}=0 
    \end{cases}
\end{equation}

\subsubsection{Test Belief Update}

Symmetrically, test beliefs are updated by treating the code population as the sensor. The likelihood roles invert ($\alpha \leftrightarrow \beta$), rewarding tests that align with high-belief code candidates:
\begin{equation}
    \label{eq:test_woe}
    \Delta_{ji}^{\text{test}} = 
    \begin{cases} 
    \ln \left( \frac{b(c_i) + \beta (1 - b(c_i))}{\alpha b(c_i) + \gamma (1 - b(c_i))} \right) & \text{if } d_{ij}=1 \\
    \ln \left( \frac{(1-\beta)(1 - b(c_i))}{(1-\alpha)b(c_i) + (1-\gamma)(1 - b(c_i))} \right) & \text{if } d_{ij}=0 
    \end{cases}
\end{equation}

\subsubsection{Analysis of Credibility Thresholds}
An analysis of the update mechanisms in Eq.~\ref{eq:code_woe} and Eq.~\ref{eq:test_woe} reveals a critical property of BACE. For a ``pass'' observation ($d_{ij}=1$) to yield a positive belief update ($\Delta > 0$) for a test $t_j$, and conversely for a ``fail'' to yield a negative update, the interacting code $c_i$ must satisfy the following inequality:
\begin{equation}
\label{eq:threshold}
b(c_i)(1 - \alpha - \beta + \gamma) > \gamma - \beta
\end{equation}
Because the term $(1 - \alpha - \beta + \gamma)$ is generally positive under realistic noise models, this condition establishes a \textbf{credibility threshold} that the interactor must exceed for the update to behave intuitively. If $b(c_i)$ falls below this threshold, the update logic inverts: a ``pass'' decreases belief in the test, while a ``fail'' increases it. This counterintuitive property is essential for system robustness, as it ensures that tests are penalized for passing low-belief (likely incorrect) code. A symmetric property applies when updating code beliefs based on the credibility of test cases, where $\alpha$ and $\beta$ are interchanged in Eq.\ref{eq:threshold}.

\subsection{Anchoring}
\label{subsec:anchoring}
To ground the co-evolutionary process, we utilize an anchor set $\mathcal{T}_{anchor}$ consisting of high-fidelity priors ($b(t_{anchor}) \approx 1$). These anchors are defined strictly as the set of public input/output examples provided in the problem specification $\mathcal{S}$. For instance, all problems in the LiveCodeBench dataset include such input/output examples. Anchoring induces an asymmetric evidence signal for the code population $\mathcal{C}$ where code that passes anchors gains immediate credibility, restricted only by the accidental pass rate $\beta$ and code that fails anchors receiving an infinite penalty. 

The anchors $\mathcal{T}_{anc}$ are always preserved in the test population and their beliefs $b(t_{anc})$ are never updated, serving as immutable references for the entire evolutionary search.

\subsubsection{Order of Belief Updates}
\label{subsubsec:orderofupdates}
Within each generation, BACE first updates the code based on Anchors to identify credible code candidates, then updates tests effectively auditing them by the most credible code candidates, and finally updates the code on test priors. Crucially, this step utilizes the test beliefs from the \textit{start} of the generation (priors) to prevent unstable feedback loops and ensure that test-driven evidence is properly dampened by their initial uncertainty. This order stabilizes the co-evolutionary dynamics and bootstrap belief propagation. 


\section{The BACE Algorithm}

Algorithm~\ref{alg:bace} presents the high-level co-evolutionary process of BACE. The process begins by initializing both the code ($\mathcal{C}$) and test ($\mathcal{T}$) populations using LLMs to establish initial diversity. Public test cases ($\mathcal{T}_{anchor}$), derived from the problem specification $S$, are extracted and assigned high prior credibility ($b(t_{anchor}) \approx 1$).

In each generation, the populations are executed against one another to form the observation matrix $\mathbf{D}$. These results serve as noisy evidence for reciprocal belief updates. The framework evolves the two populations alternatively. This cycle continues for a fixed number of generations $G_{max}$, at which point the code candidate with the highest posterior probability is selected alongside the final validated test suite.

\subsection{Extracting Anchors}
The process begins by extracting the Anchor Set ($\mathcal{T}_{anchor}$) directly from the problem specification $\mathcal{S}$ as defined in Section~\ref{subsec:anchoring}. 

\subsection{Initialization of Populations}
The two populations are initialized via an LLM to establish initial diversity for both code and test populations. Unlike the anchors, these generated populations are assigned uninformative constant priors ($b(\cdot) = 0.2$) to reflect the system's initial uncertainty regarding their validity.

\begin{algorithm}[t]
\small
\SetAlgoLined
\DontPrintSemicolon
\KwIn{Problem Specification $\mathcal{S}$, Generations $G_{max}$, Configuration $\mathcal{K}$}
\KwOut{Best Code $\hat{c}$, Validated Tests $\mathcal{T}^{(final)}$}

$\mathcal{T}_{anc} \gets \text{ExtractAnchors}(\mathcal{S})$
$(\mathcal{C}^{(0)}, \mathcal{T}^{(0)}, \mathbf{b}(\mathcal{C}^{(0)}), \mathbf{b}(\mathcal{T}^{(0)})) \gets \text{Initialize}(\mathcal{S}, \mathcal{K})$\;

\For{$g \gets 0$ \KwTo $G_{max}$}{
    $\mathbf{D}^{(g)} \gets \text{Execute}(\mathcal{C}^{(g)}, \mathcal{T}_{anc} \cup \mathcal{T}^{(g)})$ 
    
    $\mathbf{b}(\mathcal{C}), \mathbf{b}(\mathcal{T}) \gets \text{BayesianUpdate}(\mathbf{D}^{(g)}, \mathbf{b}(\mathcal{C}^{(g)}), \mathbf{b}(\mathcal{T}^{(g)}))$\;

    \uIf{$g$ is \textbf{even} \textbf{and} $g < G_{max} - 1$}{
        $\mathcal{T}^{(g+1)}, \mathbf{b}(\mathcal{T}^{(g+1)}) \gets \text{Evolve}(\mathcal{T}^{(g)}, \mathbf{b}(\mathcal{T}),  \mathcal{K})$\;
        $\mathcal{C}^{(g+1)}, \mathbf{b}(\mathcal{C}^{(g+1)}) \gets \mathcal{C}^{(g)}, \mathbf{b}(\mathcal{C})$\;
    }
    \uElseIf{$g$ is \textbf{odd} \textbf{and} $g < G_{max} - 1$}{
        $\mathcal{C}^{(g+1)}, \mathbf{b}(\mathcal{C}^{(g+1)}) \gets \text{Evolve}(\mathcal{C}^{(g)}, \mathbf{b}(\mathcal{C}),    \mathcal{K})$\;
        $\mathcal{T}^{(g+1)}, \mathbf{b}(\mathcal{T}^{(g+1)}) \gets \mathcal{T}^{(g)}, \mathbf{b}(\mathcal{T})$\;
    }
}
$\hat{c} \gets \arg\max_{c_i \in \mathcal{C}} b(c_i)$\;
\Return $\hat{c}, \mathcal{T}^{(final)}$\;
\caption{BACE: Bayesian Anchoring and Co- Evolution}
\label{alg:bace}
\end{algorithm}

\subsubsection{Code Initialization ($\mathcal{C}^{(0)}$)}

To establish a diverse starting manifold, we prompt the LLM to generate a batch of $N_a$ distinct algorithmic approaches for the specification $\mathcal{S}$ in a single response. This batch generation process is repeated $N_s$ times to populate the initial set of size $N_{init} = N_a \times N_s$.

\subsubsection{Test Initialization ($\mathcal{T^{(0)}}$)}
The initial test suite is seeded with $M_{init}$ unit tests generated to cover basic functionality, edge cases, and large-scale tests as in AgentCoder~\cite{huang_agentcoder_2024}. 

\subsection{Execution}
In every generation, the system executes the current code population $\mathcal{C}^{(g)}$ against the union of the evolved test suite $\mathcal{T}^{(g)}$ and the anchor set $\mathcal{T}_{anchor}$. This produces an observation matrix $D^{(g)}$, where each entry $d_{ij}$ represents the binary outcome (Pass/Fail) of code $c_i$ executing test $t_j$.

\subsection{Bayesian Updates}
Using the observation matrix $\mathbf{D}^{(g)}$, the system updates the belief distributions $\mathbf{b}(\mathcal{C})$ and $\mathbf{b}(\mathcal{T})$ as described in Section~\ref{subsubsec:orderofupdates}. 

\subsection{Alternating Evolution}
To mitigate the instability inherent in co-evolution (e.g., "red queen" dynamics where populations cycle without progress~\cite{cliff_tracking_1995}), BACE employs an alternating evolution strategy. In even generations, the code population is frozen while the test suite $\mathcal{T}^{(g+1)}$ evolves, while in odd generations, the test population is frozen while the code population $\mathcal{C}^{(g+1)}$ evolves. This stabilizes the learning signal, allowing one population to adapt to the stable pressure provided by the other.

\subsubsection{Elite Selection}
\label{subsubsec:elitism}
We preserve high-belief, diverse individuals across generations through a diversity-preserving elitism based on behavioral vectors. We employ distinct strategies for code and tests.

\begin{algorithm}[t]
\small
\SetAlgoLined
\DontPrintSemicolon
\KwIn{Population $\mathcal{P}$, Belief Vector $\mathbf{b}$, Config $K$}
\KwOut{New Population $\mathcal{P}'$, New Belief Vector $\mathbf{b}'$}

$\mathcal{P}_{elite}, \mathbf{b}_{elite} \gets \text{SelectElites}(\mathcal{P}, \mathbf{b}, K.\text{elitism\_rate})$\;

$N_{target} \gets \lceil |\mathcal{P}| \cdot K.\text{offspring\_rate} \rceil$\;
$\mathcal{P}_{off}, \mathbf{b}_{off} \gets \emptyset, \emptyset$\;

\While{$|\mathcal{P}_{off}| < N_{target}$}{
    $op \gets \text{SelectOperator}(K.\text{ops\_rate})$\;
    $parents \gets \text{SelectParents}(\mathcal{P}, \mathbf{b}, k=\text{arity}(op))$\;
    
    $child \gets \text{ApplyOp}(op, parents)$\;
    
    $b_{child} \gets \min_{p \in parents} \mathbf{b}(p)$\;
    
    $\mathcal{P}_{off} \gets \mathcal{P}_{off} \cup \{child\}$\;
    $\mathbf{b}_{off} \gets \mathbf{b}_{off} \cup \{b_{child}\}$\;
}

\Return $(\mathcal{P}_{elite} \cup \mathcal{P}_{off}), (\mathbf{b}_{elite} \cup \mathbf{b}_{off})$\;
\caption{BACE: Population Evolution Strategy}
\label{alg:evolve}
\end{algorithm}

\begin{figure}
\centering
\includegraphics[width=0.8\columnwidth]{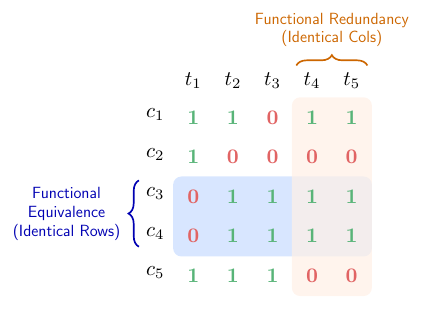}
\Description{A 5x5 matrix where rows are labeled c1 through c5 and columns t1 through t5. Cells contain pass/fail binary data. Rows c3 and c4 are highlighted in blue as functionally equivalent. Columns t4 and t5 are highlighted in orange as functionally redundant. The intersection of the highlights is a darker mixed color.}
\caption{Functional Equivalence (Blue): Candidates $c_3$ and $c_4$ produce identical pass/fail vector across all tests. 
Functional Redundancy (Orange): Tests $t_4$ and $t_5$ induce identical pass/fail vector across all code candidates.}
\label{fig:interaction_matrix}
\end{figure}

\paragraph{Code Elitism}
To prevent premature convergence, we construct the elite set $\mathcal{C}_{elite}$ via a dual-objective strategy that balances exploitation and diversity. First, we select the top-$k$ individuals with the highest posterior belief $b(\cdot)$ (Exploitation). Second, we select the highest-probability representative from every functional equivalence group identified in the population (Diversity). Following the methodology of CodeT~\cite{chen_codet_2022}, we define two candidate solutions $c_a, c_b \in \mathcal{C}$ as \textit{functionally equivalent} with respect to the current test suite if their behavior vector is identical across all tests. Formally, this corresponds to identical rows in the observation matrix $\mathbf{D}$ (as visualized in Figure~\ref{fig:interaction_matrix}) such that $\mathbf{D}_{a, \cdot} = \mathbf{D}_{b, \cdot} \iff \forall j, d_{aj} = d_{bj}$.

This clustering ensures that distinct strategic niches are preserved even if their current probability is marginally lower than the dominant mode. If the combined elite set leaves insufficient room for the required offspring quota (defined by $K.\text{offspring\_rate}$), we prune the lowest-probability individuals from this combined set until the limit is met.

\paragraph{Test Elitism}
For the test population, elitism prioritizes discriminative efficiency over belief. We define two tests $t_a, t_b \in \mathcal{T}$ as \textit{functionally redundant} if they induce identical behavior across the current code population. Formally, this corresponds to identical columns in the observation matrix $\mathbf{D}$ (as visualized in Figure~\ref{fig:interaction_matrix}) such that $ \mathbf{D}{\cdot, a} = \mathbf{D}{\cdot, b} \quad \iff \quad \forall i, d_{ia} = d_{ib}$.

Such redundancy indicates that $t_k$ offers no unique information gain over $t_j$ regarding the correctness of the current candidates. To maintain a minimal efficient basis for the belief updates, we cluster these redundant groups and retain only the single representative with the highest posterior belief $b(t)$. This effectively compresses the test suite into a set of orthogonal sensors. 

\subsubsection{Parent Selection} 
We utilize a fitness-proportionate selection (Roulette Wheel Selection) over the belief distributions. The probability of selecting an individual $i$ is directly proportional to its normalized belief score $b(i)$. The selection cardinality is operator-dependent: we sample single parents for unary operations and pairs for binary operations, applying this mechanism symmetrically across both populations.

\subsubsection{Applying Operations}
Once parents are selected, we generate offspring by applying a specific evolutionary operator, chosen stochastically at each step based on the operation rate. The selected operator dictates both the number of parents required (arity) and the specific LLM prompting strategy used to synthesize the child. For the specific details of these operators, see Section~\ref{sec:pop_and_ops}.

\subsubsection{Offspring's Belief Assignment}
Assuming the child retains the validity of its lineage until proven otherwise, we initialize the offspring's belief as the minimum of its parents' beliefs: $b_{child} = \min_{p \in \pi} b(p)$. If the offspring introduces a regression, it will fail an anchor test, triggering a catastrophic penalty ($\Delta \to -\infty$). Conversely, if it preserves functionality, it maintains a high enough belief to survive the selection cut. 

\subsection{Convergence and MAP Estimation}
After $G_{max}$ generations, the algorithm terminates and extracts the Maximum A Posteriori estimate from the final code population:

\begin{equation}
\hat{c} = \arg\max_{c_i \in C^{(final)}} b(c_i)
\end{equation}

In the event of a tie among candidates with identical maximum beliefs, we select the most recently generated offspring. 


\section{Populations and Operations}
\label{sec:pop_and_ops}

In BACE, the evolving test population is partitioned into two semantic categories: Unit Tests ($\mathcal{T}_{unit}$) and Differential Tests ($\mathcal{T}_{diff}$), while the Anchor Population ($\mathcal{T}_{anc}$) remains immutable. Unlike standard genetic algorithms that rely on stochastic syntactic mutations, BACE employs Informed Variation Operators via LLMs. 
Figure~\ref{fig:evolutionary_lineage} provides a visualization of how these operators interact with the populations to produce a candidate solution. For each operation, the LLM executes the task using a prompt conditioned on the problem specification (see Appendix for full prompt templates). 

\subsection{Code Population Operators \texorpdfstring{($\mathcal{C}$)}{C}}

\begin{description}
    \item[Semantic Crossover $(c_a \times c_b \rightarrow c')$:] Given two parent implementations, the LLM integrates the most effective logic from both parents while resolving semantic conflicts.
    
    \item[Debug $(c, \mathcal{T}_{fail} \rightarrow c')$:] This operator targets correctness. The LLM acts on a parent $c$, utilizing a context window containing a subset of failing tests $\mathcal{T}_{fail} \subset \mathcal{T}$ and their associated execution traces. To prioritize valid feedback, $\mathcal{T}_{fail}$ is sampled via rank selection based on test belief scores $b(t_j)$. The model is instructed to repair the logic to satisfy the violated constraints.
    
    \item[Re-implement $(c \rightarrow c')$:] To escape local optima, the LLM is instructed to rewrite $c$ using a fundamentally different algorithmic approach or syntactic structure, while strictly preserving its functional intent relative to $\mathcal{S}$.
\end{description}

\subsection{Unit Tests \texorpdfstring{($\mathcal{T}_{unit}$)}{(T\_unit)}}
Following operators focus on expanding the test suite's discriminative power and coverage:

\begin{description}
    \item[Discriminate] $(t \mid \{c_a, c_b\} \rightarrow t')$: Given a target code pair $\{c_a, c_b\}$ and a test $t$: if both pass $t$, the LLM is instructed to write a new test $t'$ that exposes a latent bug in the weaker candidate. If $t$ discriminates (one pass, one fail), the operator refines $t$ to better target the specific failure. If both fail, $t'$ is refined to distinguish between their failure modes.
    
    \item[Complementary Crossover $(t_a \times t_b \rightarrow t')$:] Conditioned on two parent tests $\{t_a, t_b\}$, the LLM identifies semantic gaps in their combined coverage. It synthesizes an offspring $t'$ targeting this latent "negative space," ensuring the evolution explores constraints orthogonal to existing high-fitness tests.
    
    \item[Edge Case Generation $(t \rightarrow t')$:] The LLM validates a test $t$. If $t$ is incorrect relative to $\mathcal{S}$, it is repaired. If valid, the LLM generates a variant covering a distinct input scenario or boundary condition not covered by the parent.
\end{description}

\subsection{Differential Tests \texorpdfstring{($\mathcal{T}_{diff}$)}{T\_diff}}
The objective of $\mathcal{T}_{diff}$ is to generate a distinguishing input $t'$ that induces divergent outputs between the candidates, i.e., $\mathcal{O}(c_a, t') \neq \mathcal{O}(c_b, t')$. 
$\mathcal{T}_{diff}^{(0)} = \emptyset$, and tests are added dynamically only when equivalence clusters are detected. We define a single operator for this purpose:

\begin{description}
    \item[Divergence Discovery] $(\{c_a, c_b, \mathcal{T}_{pass}\} \rightarrow \{t'_k\} \subset \mathcal{T}_{diff})$: To split a cluster of candidates $\{c_a, c_b\}$ that exhibit identical behavior vectors, the LLM is prompted to synthesize a Python input generator script. Crucially, we condition this prompt on the subset of tests $\mathcal{T}_{pass} \subseteq \mathcal{T}$ that both candidates currently pass. This context allows the model to infer the shared logic and target edge cases where the implementations likely differ. This methodology is adapted from the Mokav framework~\cite{etemadi_mokav_2025}. The script produces stochastic inputs $i_k$, on which we execute both candidates. If the outputs diverge ($\mathcal{O}_a \neq \mathcal{O}_b$), we capture the specific input $i_k$ and both divergent outputs as two separate test candidates: $t'_a = (i_k, \mathcal{O}_a)$ and $t'_b = (i_k, \mathcal{O}_b)$. While one of these tests is necessarily incorrect relative to $\mathcal{S}$, the Bayesian reciprocity mechanism (Section 3.2) will naturally filter the erroneous case.
\end{description}
 
\begin{table*}[t]
  \centering
  \small 
  \caption{Comparison of Pass@1 (\%) performance on LiveCodeBench v6 (Post-March 2025) across problem difficulty levels. Results are reported as Average$_{\pm \text{Std}}$ over 3 runs.}
  \label{tab:main_results}
  \setlength{\tabcolsep}{4pt} 
  \begin{tabular}{l cccc c cccc c cccc}
    \toprule
    & \multicolumn{4}{c}{\textbf{GPT-5-Mini}} & & \multicolumn{4}{c}{\textbf{Qwen2.5-Coder:7b}} & & \multicolumn{4}{c}{\textbf{GPT-OSS-120b}} \\
    \cmidrule(lr){2-5} \cmidrule(lr){7-10} \cmidrule(lr){12-15}
    \textbf{Method} & Easy & Med & Hard & Overall & & Easy & Med & Hard & Overall & & Easy & Med & Hard & Overall \\
    \midrule
    Direct     & $83.3_{\pm 5.6}$ & $48.0_{\pm 10.6}$ & $24.3_{\pm 2.7}$ & $45.0_{\pm 4.5}$ & & $38.9_{\pm 5.6}$ & $5.3_{\pm 2.3}$  & $1.8_{\pm 1.6}$  & $11.3_{\pm 1.3}$ & & $96.3_{\pm 3.2}$ & $76.0_{\pm 4.0}$ & $26.1_{\pm 1.6}$ & $57.5_{\pm 1.3}$ \\
    \addlinespace
    AgentCoder & $92.6_{\pm 3.2}$ & $65.3_{\pm 14.1}$ & $35.1_{\pm 2.7}$ & $57.5_{\pm 4.5}$ & & $38.9_{\pm 5.6}$ & $4.0_{\pm 4.0}$  & $2.7_{\pm 2.7}$  & $11.3_{\pm 3.3}$ & & $79.6_{\pm 12.8}$& $69.3_{\pm 2.3}$ & $27.0_{\pm 4.7}$ & $52.1_{\pm 5.8}$ \\
    \addlinespace
    MapCoder   & $85.2_{\pm 3.2}$ & $68.0_{\pm 4.0}$  & $36.9_{\pm 8.3}$ & $57.5_{\pm 4.3}$ & & --               & --               & --               & --               & & $46.3_{\pm 11.6}$& $38.7_{\pm 9.2}$ & $18.0_{\pm 1.6}$ & $30.8_{\pm 5.2}$ \\
    \addlinespace
    CodeSIM    & $92.6_{\pm 3.2}$ & $77.3_{\pm 8.3}$  & $41.4_{\pm 9.5}$ & $64.2_{\pm 7.5}$ & & $64.8_{\pm 8.5}$ & $20.0_{\pm 4.0}$ & $7.2_{\pm 1.6}$  & $24.2_{\pm 2.6}$ & & $90.7_{\pm 6.4}$ & $82.7_{\pm 6.1}$ & $46.0_{\pm 7.2}$ & $67.5_{\pm 6.5}$ \\
    \midrule
    \textbf{BACE (Ours)} & \textbf{96.3}$_{\pm 3.2}$ & \textbf{78.7}$_{\pm 6.1}$ & \textbf{44.1}$_{\pm 10.2}$ & \textbf{66.7}$_{\pm 4.4}$ & & \textbf{72.2}$_{\pm 0.0}$ & \textbf{26.7}$_{\pm 2.3}$ & \textbf{10.8}$_{\pm 5.4}$ & \textbf{29.6}$_{\pm 1.9}$ & & \textbf{100.0}$_{\pm 0.0}$ & \textbf{86.7}$_{\pm 2.3}$ & \textbf{49.6}$_{\pm 5.6}$ & \textbf{72.5}$_{\pm 2.5}$ \\
    \bottomrule
  \end{tabular}
\end{table*}
\section{Experiments and Results}
\subsection{Experimental Setup}

\subsubsection{Dataset}
We evaluate on LiveCodeBench v6, utilizing an 80-problem subset (18 Easy, 25 Medium, 37 Hard) published post-March 2025. The choice benchmark avoids contamination, and selecting a recent filter date allows us to use new models.

\subsubsection{Language Models} 
We evaluate BACE across three distinct language models: the proprietary \textbf{GPT-5-Mini}, alongside two open-weight models spanning different scales, \textbf{Qwen2.5-Coder-7b} and \textbf{GPT-OSS-120b}. Notably, we selected models having a reported pre-training knowledge cutoff before LiveCodeBench v6 filter date. The following configurations were used for the models: for GPT-5-Mini, we set the \texttt{reasoning\_effort} to \texttt{minimal}, other parameters such as temperature are not configurable for this model. For GPT-OSS-120b, we adopt the recommended settings of \texttt{temperature}=1.0, \texttt{top\_p}=1.0, and we set the \texttt{reasoning\_effort} to \texttt{low} reasoning. For Qwen2.5-Coder-7b, we set the \texttt{temperature} to 0.7.

\subsubsection{State-of-the-Art and Baselines}
CodeSIM~\cite{islam_codesim_2025} paper and AgentCoder~\cite{huang_agentcoder_2024} 2024 updated paper have reported the state-of-the-art results. We compared BACE against the top two systems identified in the CodeSIM paper results (MapCoder~\cite{islam_mapcoder_2024}, and CodeSIM~\cite{islam_codesim_2025}), AgentCoder~\cite{huang_agentcoder_2024}, and against direct prompting as a baseline. 

\subsection{BACE Configuration}
To ensure reproducibility, we specify the evolutionary parameters employed in our experiments. We run the co-evolution for $G_{max}=10$ generations. The code population is initialized with $N_{init}=10$ candidates and capped at a maximum size of $N_{max}=15$, utilizing an elitism rate of 0.3 and an offspring generation rate of 0.3. Code evolutionary operators are selected stochastically, prioritizing Debugging ($p=0.6$) over Rewriting ($p=0.2$) and Semantic Crossover ($p=0.2$). The unit test population is initialized with a fixed size of $M=20$ candidates. Test evolution operators are applied with probabilities of 0.5 for Discrimination, 0.3 for Edge Case Generation, and 0.2 for Complementary Crossover. Additionally, for differential testing, the system generates 10 tests (derived from 5 diverging inputs) for every identified divergent pair.

\subsubsection{Hyperparameters and Calibration}
We adjust these hyperparameters based on the test source and the initial beliefs, setting the initial belief for both code and tests to $b_{init}=0.2$. For public anchor tests ($\mathcal{T}_{anc}$), since these are ground truth, we set $\alpha=\gamma=0$, while setting $\beta=0.2$ to account for the limited discriminative power of trivial anchors. Conversely, for evolved tests ($\mathcal{T}_{unit}$ and $\mathcal{T}_{diff}$), we selected parameters of $\alpha=0.1$, $\beta=0.2$, and $\gamma=0.25$ by running a parameter sweep to check for convergence. 

\subsection{Implementation Details}
We run all experiments on a 16-core AMD EPYC 7282 server (64GB RAM), utilizing a local NVIDIA RTX A4000 (16GB VRAM) for Qwen2.5-Coder-7b, and the Groq and OpenAI APIs for GPT-OSS-120b and GPT-5-Mini. Source code and experimental scripts are available on GitHub at \url{https://github.com/wso2-incubator/BACE} and archived on Zenodo (\href{https://doi.org/10.5281/zenodo.19508306}{DOI: 10.5281/zenodo.19508306}).
\subsection{Results}

Table~\ref{tab:main_results} summarizes the Pass@1 performance of BACE against state-of-the-art baselines across the  GPT-5-Mini, Qwen2.5-Coder-7b and GPT-OSS-120b, reporting the average and standard deviation over three runs.

\newcommand{\best}[1]{\textbf{#1}}

As detailed in Table~\ref{tab:main_results}, BACE consistently outperforms all baselines across every difficulty level for each evaluated model. Overall, BACE establishes a new state-of-the-art, achieving absolute margins of 5.0\% (72.5\% vs. 67.5\%) on GPT-OSS-120b, 2.5\% (66.7\% vs. 64.2\%) on GPT-5-Mini, and 5.4\% (29.6\% vs. 24.2\%) on Qwen2.5-Coder-7b over CodeSIM.

Running with GPT-OSS-120b and Qwen2.5-Coder-7b, MapCoder frequently failed to adhere to the structural schema required by the prompt, resulting in XML parsing errors and code syntax violations. As this limitation is exacerbated in the smaller Qwen2.5-Coder-7b model, MapCoder was omitted from Qwen2.5 evaluations.

\subsection{Ablation Studies}

To isolate the impact of our co-evolutionary approach, we conduct an ablation study on the Hard problem subset ($N=37$) using GPT-OSS-120b. We evaluate six configurations that trace the progression from single-solution generation to full co-evolution, reporting the average over three runs: (1) direct prompting (a single-solution baseline), (2) population-based filtered sampling (no evolution), where a candidate is sampled from an initial generated population after filtering with public anchors ($\mathcal{T}_{anc}$), (3) population-based filtered sampling where the candidate passing $\mathcal{T}_{anc}$ and the highest number of static unit tests ($\mathcal{T}_{unit}$) is selected, (4) code evolution guided solely by public anchors, (5) code evolution guided by anchors and static unit tests, and (6) full BACE co-evolution.

As detailed in Table~\ref{tab:ablation}, relying on a single-solution strategy via direct prompting yields the lowest performance at 26.1\%. Moving to a population-based strategy provides an immediate benefit: extracting a solution via static filtered sampling improves performance to 29.7\% with anchors alone, and establishes a further improvement reaching 33.3\% when incorporating static unit tests as a ranking signal over the population. Transitioning from static population sampling to code population evolution provides a substantial jump, reaching 41.4\% with anchors and 44.1\% with static unit tests. By evolving tests alongside the code population, the full BACE framework achieve the highest Pass@1 score of 49.6\%.

\begin{table}[h]
  \centering
  \small
  \caption{Ablation study evaluating the impact of test case strategies on the Hard problem subset ($N=37$) using GPT-OSS-120b. Performance is measured as Pass@1 (\%) reported as Average$_{\pm \text{Std}}$ over 3 runs.}
  \label{tab:ablation}
  \begin{tabular}{@{} l c @{}}
    \toprule
    \textbf{Ablation Configuration} & \textbf{Pass@1 (\%)} \\
    \midrule
    Direct Prompting (Single Solution) & $26.1_{\pm 1.6}$ \\
    \addlinespace
    \textit{Population Sampling (No Evolution):} & \\
    \quad Filtered Sampling ($\mathcal{T}_{anc}$ Only) & $29.7_{\pm 0.0}$ \\
    \quad Filtered Sampling ($\mathcal{T}_{anc}$ + Static $\mathcal{T}_{unit}$) & $33.3_{\pm 5.6}$ \\
    \addlinespace
    \textit{Population Evolution:} & \\
    \quad Code Evolution ($\mathcal{T}_{anc}$ Only) & $41.4_{\pm 4.1}$ \\
    \quad Code Evolution ($\mathcal{T}_{anc}$ + Static $\mathcal{T}_{unit}$) & $44.1_{\pm 7.8}$ \\
    \midrule
    \textbf{Full BACE (Co-Evolution)} & \textbf{49.6}$_{\pm 5.6}$ \\
    \bottomrule
  \end{tabular}
\end{table}
\section{Discussion and Lessons Learned}
Our results on LiveCodeBench v6 (post-March 2025) show evidence that BACE surpasses state-of-the-art performance with proprietary models as well as open weight small and large models across different difficulty levels of problem specifications. This performance supports our central hypothesis: that in the absence of a formal oracle, LLM-generated tests provide a strong learning signal when modeled as noisy sensors within a probabilistic framework.

\subsection{Reclaiming the Signal from Generated Tests}

AgentCoder occasionally underperforms even the Direct baseline (52.1\% vs 57.5\% overall with GPT-OSS-120b). We attribute this to the 'False Negative Rejection' failure mode detailed in the introduction, where faulty tests erroneously degrade valid solutions. As seen in the ablation results( Table~\ref{tab:ablation}), population-based methods offer a higher likelihood that the original valid logic will remain within the broader population, even if a specific candidate undergoes such a destructive update. 

However, population-based methods too are susceptible to drift. Ideally, correct tests serve to preserve accurate solutions while eliminating incorrect ones. Yet, the presence of faulty tests diminishes the fitness of valid solutions, prompting destructive updates that favor candidates aligning with the errors. Consequently, these incorrect solutions might gain fitness from the faulty tests, creating a feedback loop that drives the population away from the initial specification. 

Within BACE, co-evolutionary drift requires an incorrect solution to satisfy two simultaneous conditions: it must pass the anchors ($T_{anc}$) and it must pass high-belief tests. Crucially, these tests achieve high-belief only after being validated by the positively anchored solutions. However, the likelihood of an incorrect solution meeting both criteria while still being wrong is much smaller. In other words, as observed by CodeT~\cite{chen_codet_2022}, it is easier for correct solutions and tests to agree than wrong solutions and wrong tests to agree as there are many ways to be wrong than ways to be correct. Furthermore, the moment an incorrect solution fails an anchor, Bayesian updates propagate this negative signal, reducing the belief in any faulty tests that support the incorrect code. This mechanism effectively breaks the drift feedback loop, resulting in faster convergence. 
\subsection{Importance of Population Diversity}
The success of BACE is rooted in an understanding of solution lifecycles. By tracking the lifecycle of individual code and test identities from generation to extinction, we observed that standard elitism policies (e.g: Global Elitism) can be harmful in this domain. Before BACE, we initially explored a naive belief-based global elitism strategy that retained individuals with the firmest beliefs across generations, leading to co-evolutionary drift. 

To avoid the system settling on trivial assertions and solutions, we changed the test elitism policy to also favor the tests that discriminate using an entropy based discrimination score. However, we noticed that tests that induced different behavior in the code population were killed off, either due to lower discrimination scores or lower beliefs. This happens because scalar metrics (i.e discrimination, failure rate) could not adequately capture behavioral vectors, discarding unique tests that target different behaviors.

We believe diversity is the key workhorse behind BACE, and it employs several approaches to preserve diversity. 

First, BACE clusters individuals in both populations based on their execution behavior vectors as detailed in Section~\ref{subsubsec:elitism}. In the code population, this preserves distinct strategic niches, even retaining the sub-optimal solutions that fail anchors which can be then used by the Bayesian model to provide negative signals for the belief updates. Similarly, this prevents the test suite from filling up with redundant sensors that offer no new information gain. 

Second, unlike traditional software engineering, where differential testing is used as a bug-finder, BACE uses it for diversity retention. Although differential tests do not generate new code diversity directly, they serve as discriminators that partition functional equivalence groups. The differential test splits the group, compelling the code elitism mechanism to select and retain better representatives from each newly formed cluster. This ensures that there is sufficient diversity for evolution to progress. 

\subsection{Anchored Dual Agreement}
We observed from execution traces that once a candidate solution satisfied the public anchors, BACE consistently propagated its logic to future generations. This persistence occurred regardless of whether the solution appeared in the initial population or evolved later, effectively withstanding the abundance of incorrect candidates. We hypothesize that this persistence is driven by the same "dual agreement" phenomenon observed in CodeT~\cite{chen_codet_2022}.

However, in BACE, this agreement is strictly regulated by the anchoring process. Unlike unconstrained consensus, which can drift into spurious agreement, our model restricts unfavorable belief propagation: if a consensus cluster fails the public anchors, it hurts the beliefs of both the code and the agreeing tests. It is only when the consensus cluster is positively anchored (i.e., satisfying the ground truth) that the dual agreement mechanism effectively drives the beliefs of both populations, amplifying the signal of the valid minority and propagating their logic to the offspring.

\subsection{Future Directions}
The modular architecture of BACE establishes a flexible foundation for future research and open source development, particularly due to the decoupling of evolutionary logic from LLM-driven operators. For instance, inspired by CodeSIM's simulation agent, future "Debug" operators could leverage runtime logs to provide richer debugging signals. Furthermore, in addition to differential testing, we plan to explore how we can incorporate the state of the art testing methodologies such as Property-Based Testing~\cite{he_use_2025} or Mutation Testing~\cite{tip_llmorpheus_2025} to rigorously challenge candidates beyond unit and differential tests. Finally, further investigating the potential for anchor-free evolution remains a priority for extending BACE to scenarios where no public input/output examples are available. 


\section{Conclusion}



The state-of-the-art in LLM-driven code generation has largely abandoned test generation for reasoning-based approaches, citing the unreliability of synthesized tests as feedback. We argue, conversely, that generated tests remain a high-value signal when modeled as noisy sensors within a Bayesian framework. BACE validates this hypothesis through a Bayesian co-evolutionary process where code and test populations reciprocally refine one another. By anchoring this search on minimal public examples, we effectively mitigate the co-evolutionary drift typical of self-validating loops. Furthermore, we utilize behavioral based elitism policy and differential testing to enforce population diversity, a mechanism that prevents the system from collapsing into redundant or trivial solutions. Empirical evaluations on LiveCodeBench v6 (Post-March) demonstrate that BACE establishes a new state-of-the-art, yielding performance gains across both propri-
etary models and open-weight models at 7B and 120B scales.

Looking forward, the modular Bayesian backbone of BACE provides a flexible foundation for future research. Specifically, it opens pathways for integrating state-of-the-art code and test generation to co-evolving populations to yield increasingly robust solutions.

\newpage

\bibliographystyle{ACM-Reference-Format}
\bibliography{references}

\end{document}